\documentclass{article}

\usepackage{titlesec}

\titleformat{\paragraph}
{\normalfont\normalsize\bfseries}{\theparagraph}{1em}{}
\titlespacing*{\paragraph}
{0pt}{3.25ex plus 1ex minus .2ex}{1.5ex plus .2ex}

\newcommand{\beginsupplement}{%
        \setcounter{table}{0}
        \renewcommand{\thetable}{S\arabic{table}}%
        \setcounter{figure}{0}
        \renewcommand{\thefigure}{S\arabic{figure}}%
     }

\PassOptionsToPackage{nonatbib}{nips_2018_latest}

\usepackage[final]{nips_2018_latest}

\usepackage{graphicx}

\usepackage{amsmath}
\usepackage[utf8]{inputenc} 
\usepackage[T1]{fontenc}    
\usepackage{hyperref}       
\usepackage{url}            
\usepackage{booktabs}       
\usepackage{amsfonts}       
\usepackage{nicefrac}       
\usepackage{microtype}      
\usepackage{color,soul}

\usepackage{placeins}
\usepackage{tikz}
\usetikzlibrary{shapes,arrows,positioning}
\usepackage{url}
\usepackage{amsmath}

\usepackage[sorting=none,backend=biber]{biblatex}
\hypersetup{
    colorlinks=true,
    linkcolor=red,
    filecolor=red,      
    urlcolor=red,
    citecolor=blue
}

\usepackage{float}
\usepackage{caption}
\usepackage{fixmath}

\title{Adjusting for Confounding in Unsupervised Latent Representations of Images}

\author{Craig A. Glastonbury\textsuperscript{\textbf{$\ast$}}\\
BenevolentAI, London (UK) \\ 
Big Data Institute, University of Oxford \\
\texttt{craig.glastonbury@benevolent.ai}
\And
Michael Ferlaino\thanks{\textbf{Joint first authors.}}\\
Big Data Institute, University of Oxford\\
\texttt{michael.ferlaino@bdi.ox.ac.uk}
\And Christoffer Nell\aa ker\textsuperscript{$\dagger$}\textsuperscript{$\ddagger$}\\
Big Data Institute, University of Oxford
\And Cecilia M. Lindgren\thanks{Joint last authors.}\,\,\,\thanks{A complete list of these authors' affiliations can be found in the Supplementary Materials.}\\
Big Data Institute, University of Oxford
}

\begin{document}
\maketitle
\begin{abstract}
Biological imaging data are often partially confounded or contain unwanted variability. Examples of such phenomena include variable lighting across microscopy image captures, stain intensity variation in histological slides, and batch effects for high throughput drug screening assays. Therefore, to develop ``fair'' models which generalise well to unseen examples, it is crucial to learn data representations that are insensitive to nuisance factors of variation. In this paper, we present a strategy based on adversarial training, capable of learning unsupervised representations invariant to confounders. As an empirical validation of our method, we use deep convolutional autoencoders to learn unbiased cellular representations from microscopy imaging.
\end{abstract}

\section{Introduction}\label{intro}
Recently, there has been growing evidence showing that machine learning systems can lead to poor generalisation due to inductive biases \cite{mcelany2017}. For instance, it has been shown that algorithms trained on biased data sets end up discriminating against minorities when allocating resources and/or opportunities - see \textit{e.g.} \cite{barocas2016big,podesta2014big}. This has led to the realisation that algorithmic decision making needs to go beyond ``simple'' minimisation of errors, in order to learn ``fair'' hypotheses.

For biomedical applications, solving cognitive tasks by learning supervised deep representations requires large amounts of labelled data. However, ground truth annotations are extremely expensive and time consuming since they require domain experts. Furthermore, the collection of high confidence human labelling is hindered by both interobserver and intraobserver variability - see \textit{e.g.} \cite{bueno2009impact,sorensen1993interobserver,kidron2017automated,al2015delayed}.

Given such considerations, in this work we develop a deep learning method capable of learning unsupervised representations of images that encode meaningful biological knowledge whilst being invariant to specified confounding variables. We enforce fairness into latent representations by exploiting adversarial training to remove ``noise signals'' induced by confounders. 

To validate our approach, we consider a high throughout microscopy imaging assay that captures morphological perturbations in cancer cell lines exposed to a compendium of drugs. Microscopy assays enable the analysis of cellular phenotypes through quantification of cell morphology. In particular, profiling of small molecules has the potential to deepen our understanding into mechanisms of action for specific compounds, thereby enabling faster drug development cycles \cite{scheeder2018machine}.


\section{Problem Statement}\label{problem_statement}
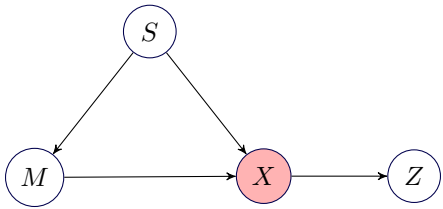
\begin{figure}[ht!]
\begin{center}
\begin{tikzpicture}[
    node distance = 14.1 mm and 10 mm,
                > = stealth',
every node/.style = {circle, draw=blue!30!black,
                    minimum size=7mm}
                    ]
\node (S) [] {$S$};
\node (X) [below right= of S, fill=red!30]    {$X$};
\node (M) [below left = of S]    {$M$};
\node (Z) [right of = X, node distance = 2cm] {$Z$};


\draw[->] (S) -- (X);
\draw[->] (S) -- (M);
\draw[->] (M) -- (X);
\draw[->] (X) -- (Z);
\end{tikzpicture}

\end{center}
\caption{\textbf{Causal relationships between factors of variation within biased image sets.} Each image $\mathbf{x}_{i}$ is drawn from a data generating distribution caused by factors $M$ and nuisance $S$. Due to inherent bias within the data, latent representations $\mathbf{z}_{i}$ are marginally dependent on the confounder $s$. }
\label{diagram}
\end{figure}

We model our learning task as follows. We are given a data set of images $\mathbf{x}_{i}\in {\rm I\!R}^{h\times w \times c}$ where $(h, w,c)$ denote height, width, and number of colour channels, respectively. Images are sampled from a data generating distribution characterised by the presence of both informative $M$ and nuisance factors $S$, with some $M$'s confounded by $S$ (Figure \ref{diagram}).

Similar to previous work using neural networks to learn fair classifiers on toy and text data \cite{louppe,zhang,controllable_invar}, we aim to learn, for each image $\mathbf{x}$ with associated nuisance factor $s$, a latent representation $\mathbf{z}\in {\rm I\!R}^{d}$ (where $d \ll h \cdot w \cdot c$), independent of $s$. First, we train a convolutional autoencoder (CAE) to learn the reconstruction $\hat{\mathbf{x}}=f(\mathbf{x}, \mathbold{\theta})$. The CAE is comprised of a cascade of two modules, where the encoder is estimating the posterior $p(\mathbf{z}\,|\, \mathbf{x})$ and the decoder is reconstructing the input image. Furthermore, we use an adversarial network $g(\textbf{z}, \mathbf{w})$ to estimate $p(s\,|\,\textbf{z})$. Given the CAE and adversarial losses $(\mathcal{L}_{\textrm{CAE}}, \mathcal{L}_{\mathrm{adv}})$, we introduce the joint objective $E$:
\begin{equation}\label{joint}
    E_{\lambda}(\mathbold{\theta}, \mathbf{w}) = \mathcal{L}_{\textrm{CAE}}(\mathbold{\theta}) -\lambda\,\mathcal{L}_{\mathrm{adv}}(\mathbold{\theta}, \mathbf{w})
\end{equation}

Therefore, deep representations which are invariant to $s$ can be learned by solving the following min/max optimisation problem \cite{louppe,zhang,controllable_invar}:
\begin{equation}\label{solution}
    \mathbold{\theta}_{*}, \mathbf{w}_{*}\,=\,\arg\,\min\limits_{\mathbold{\theta}}\,\max\limits_{\mathbf{w}}\,E_{\lambda}(\mathbold{\theta}, \mathbf{w})
\end{equation}
The architecture used for adversarial learning is visualised in Figure \ref{architecture}. All implementation details about our training protocol can be found on the GitHub repository accompanying this work (\url{https://github.com/Nellaker-group/FairUnsupervisedRepresentations}).

\section{Data}\label{dataset}
We used the BBBC021v1 image set \cite{broaddataset}, available from the Broad Bioimage Benchmark Collection \cite{ljosa2012annotated}. These data are comprised of microscopy images capturing phenotype changes in breast cancer cell lines, after being exposed to several drugs at different concentrations. Phenotypes were captured by labelling cells with DAPI, Tubulin, and Actin, thereby generating a triplet of single channel images (one per fluorescent marker) for each treatment. Within such a resource, imaged across 10 weeks (batches), a subset of treatments have been annotated with a corresponding molecular mechanism of action (MOA). We focus our attention to 92 treatments obtained after removing the two mechanisms of action (cholesterol lowering and kinase inhibitors) that appear in single batches - see \cite{broaddataset,ljosa2012annotated,ljosa2013comparison} for details.

The BBBC021v1 image set was specifically annotated as benchmark data to validate image based profiling methods and, consequently, this data set has been extensively analysed. For instance, recent works have developed deep learning approaches heavily reliant on transfer learning, in order to accurately predict MOA labels - see, \textit{e.g.} \cite{pawlowski2016automating,ando2017improving,kensert2018transfer}.

\begin{figure}[ht!]
\centering
\includegraphics[scale=0.5]{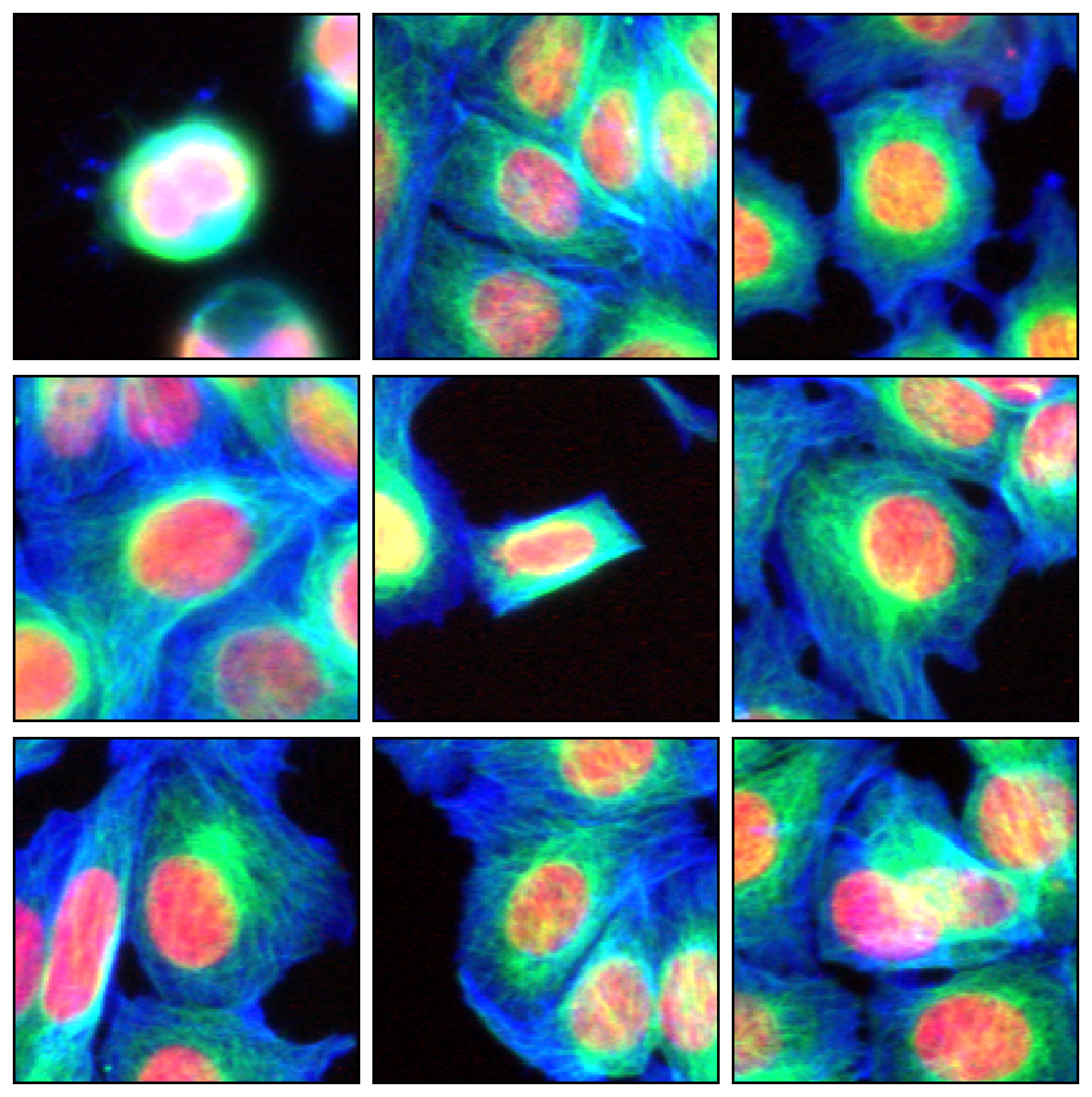}
\centering
\caption{\textbf{Examples of data instances used to develop our approach.} We trained our CAE using $128 \times 128 \times 3$ images, centred on nuclei, obtained by stacking patches from the DAPI (R), Tubulin (G), and Actin (B) channels. The ``R channel'' is activated by cell nuclei whilst the ``G channel'' highlights the characteristic Tubulin halo surrounding each nucleus.}
\label{examples}
\end{figure}

For each image in the BBBC021v1 set, we detected cell nuclei using the algorithm difference of Gaussians \cite{lindeberg1998feature} on the DAPI channel. We cropped patches of $128 \times 128$ pixels centred around each nucleus, and annotated $128 \times 128 \times 3$ images by concatenating patches from the DAPI, Tubulin, and Actin channels. Representative images used to develop our model are visualised in Figure \ref{examples}.

\section{Results}\label{results}
\begin{figure}[ht!]
\centering
\includegraphics[scale=0.6]{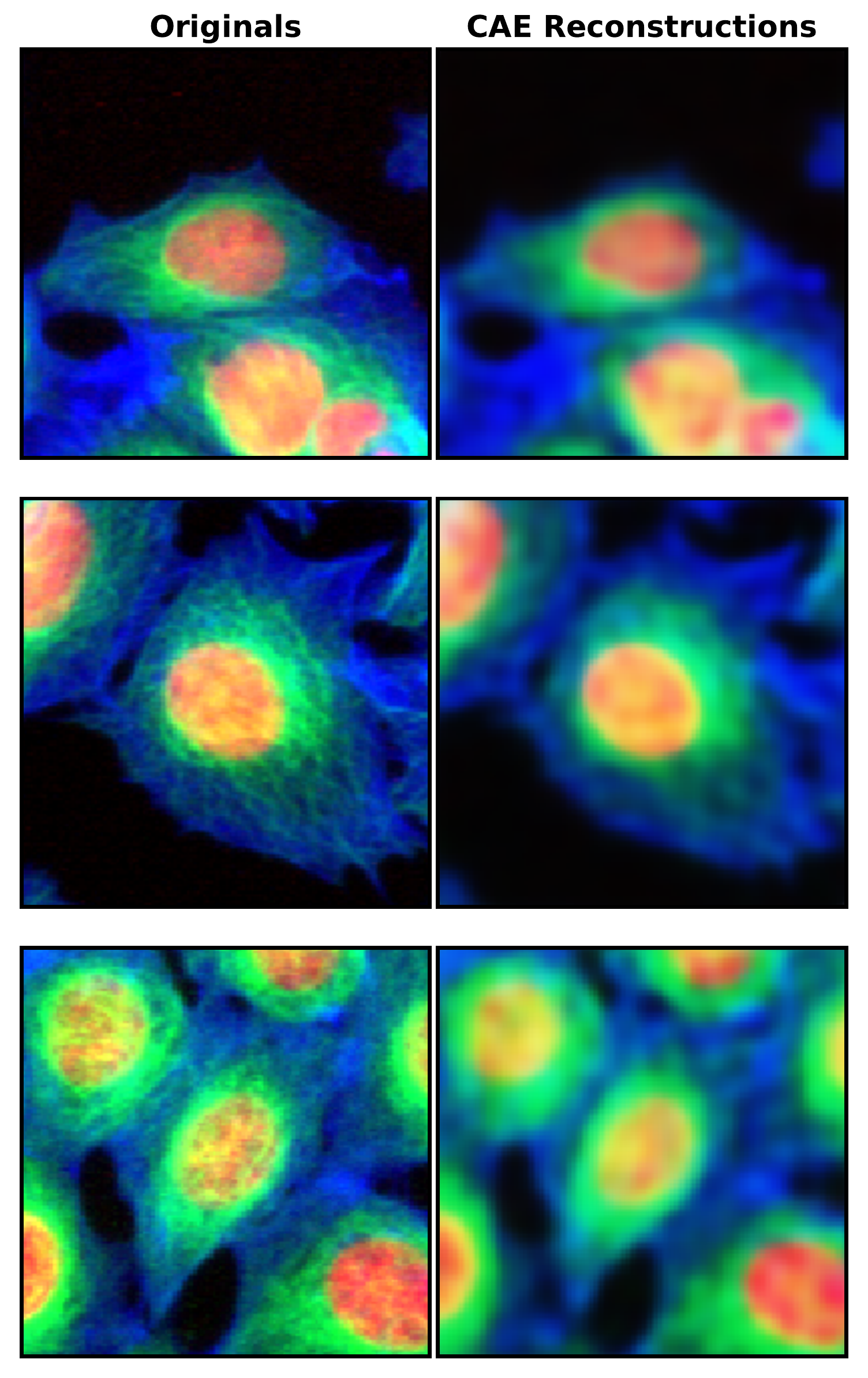}
\centering
\caption{\textbf{CAE reconstructions, of test images, before adversarial training.} Visual comparison between test images (left column), and their corresponding reconstructions (right column).}
\label{recons}
\end{figure}

Contrary to previous methods, this work is concerned with learning unsupervised representations of cells, by means of deep convolutional autoencoders. The CAE was trained with Adam \cite{kingma2014adam} for 200 epochs, achieving a training (test) loss of 0.00297 (0.00297). Furthermore, as qualitative assessment, in Figure \ref{recons} we display some (test) reconstructions. Our CAE has not only learned about the global structure of images, but it is also capable of capturing fluorescence variance across cells, as well as intracellular luminescence gradients.

\begin{figure}[ht!]
\centering
\includegraphics[scale=0.4]{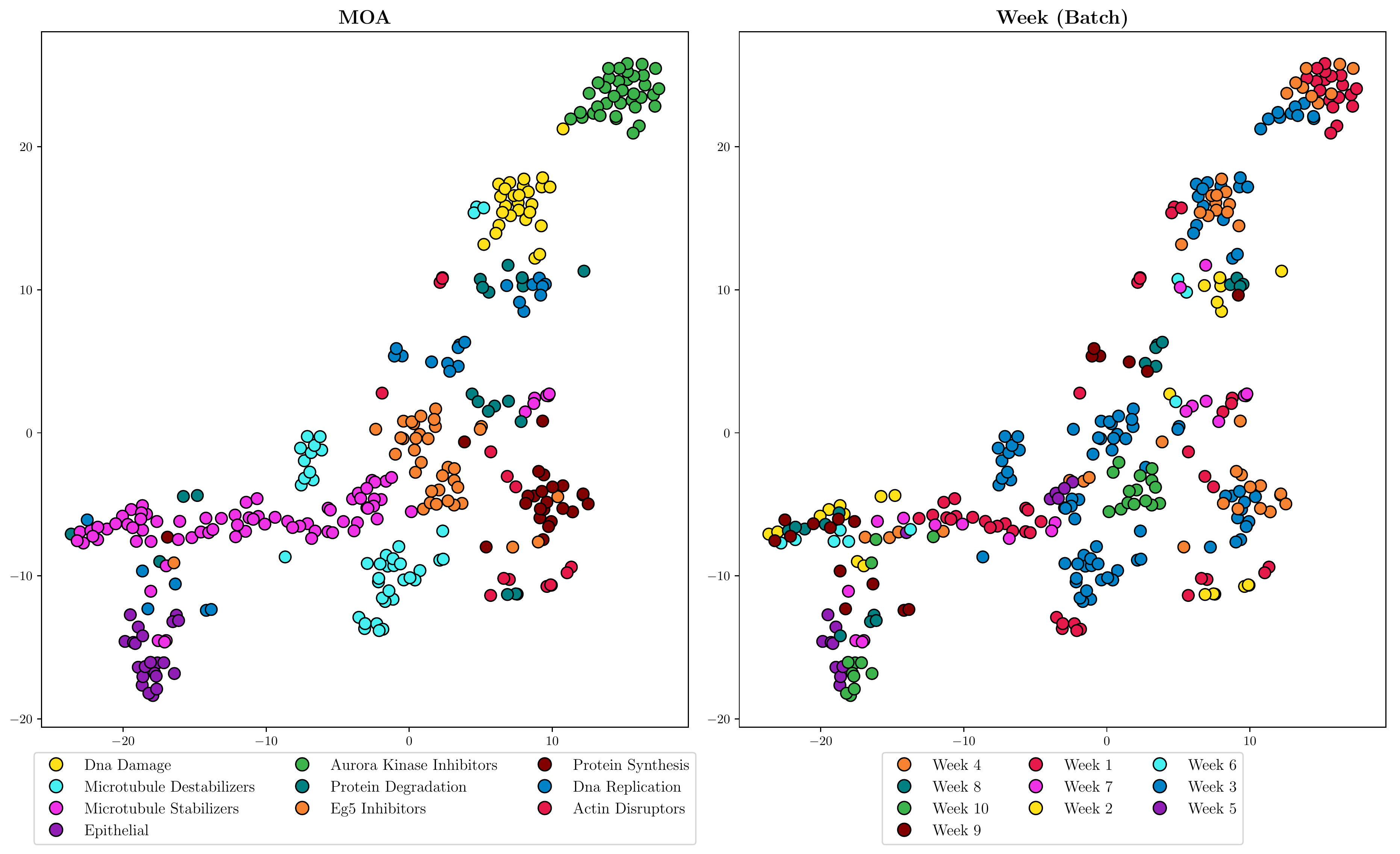}
\centering
\caption{\textbf{2D projections of CAE's latent representations, before adversarial training.} Before implementing an adversarial optimisation programme, several sets of data points (representing treatments) cluster according to their MOA label (knowledge) as well as imaging batch (confounder).}
\label{tSNE}
\end{figure}

In order to quantitatively assess whether CAE representations are capturing meaningful biological knowledge, we implemented an approach similar to previous works \cite{pawlowski2016automating,ando2017improving}. We used CAE codings as feature vectors in a three nearest neighbour (3NN) classifier to predict MOA labels for the subset of images with annotations of molecular mechanisms (92 treatments). We assessed the expressiveness of our CAE representations by implementing a leave one out cross validation protocol, and compared the results against null accuracy. As can be seen in Table \ref{tab:lambdas}, learned CAE representations capture meaningful biological knowledge as well as experimental confounding, since we were able to accurately predict MOA labels and discriminate treatments according to their batch (week). The presence of a strong batch effect has been documented in these data before, but no attempt, to the best of our knowledge, has been undertaken to precisely quantify the extent to which such nuisance factor is boosting MOA accuracies, or whether confounding knowledge can be removed from learned representations \cite{ando2017improving}. The existence of a strong confounder is also manifest in Figure \ref{tSNE}, where we display $t$SNE projections \cite{maaten2008visualizing} of CAE codings for the (92) treatments with annotated mechanism of action. Figure \ref{tSNE} shows several sets of data points which are clustered according to their MOA label as well as their imaging batch (also see \cite{ando2017improving}, Figures S1 and S2).


In this work we show that by means of adversarial training it is possible to remove confounding encoded in unsupervised cellular representations, whilst preserving biological expressiveness. Our results also represent empirical estimates that quantify to which extent the batch effect confounding is helping the MOA classification task. We used a training protocol similar to previous work \cite{louppe,controllable_invar}, by implementing a min/max optimisation that minimises the reconstruction loss whilst removing batch effects from CAE representations (Section \ref{problem_statement}). This protocol is characterised by the presence of the hyperparameter $\lambda$, where $\lambda > 1$ emphasises fairer, less biased representations, whilst resulting in higher CAE reconstruction losses. To assess the quality of our latent representations after adversarial training, we, once more, used CAE codings as feature vectors for a nearest neighbour classifier to predict MOA labels and imaging batches. We analysed how MOA and batch accuracies change across different values of $\lambda$, and the results of our experimentation are collected in Table \ref{tab:lambdas}. The adversarial programme was successful in removing the batch effect confounder, whilst preserving meaningful biological knowledge helpful for the MOA classification task. Higher values of $\lambda$ result in reductions of the CAE codings' power to discriminate batches. At $\lambda = 50$, our CAE codings are uninformative for the batch effect confounder, whilst still capable of successfully inferring MOA labels about $4\times$ better than random chance. Furthermore, the decrease in MOA accuracies represent additional evidence for the strength of the confounder present in the BBBC021v1 image set.
\begin{table}[ht!]
\vspace{+0.2cm}
\centering
\begin{tabular}{@{}cccc@{}}
\,\,\,\textbf{3NN} & \textbf{MOA Accuracy (FC)} & \textbf{Batch Accuracy (FC)} \,\,\, \\ \midrule \midrule
$\lambda = 0$              &  84.783\% $(5.6 \times)$   &   64.130\% $(2 \times)$  \vspace{+0.1cm} \\
\hline \vspace{-0.3cm}\\
$\lambda = 1$              &  59.783\% $(3.9 \times)$   &   42.391\% $(1.3 \times)$  \vspace{+0.1cm} \\
$\mathbold{\lambda}=\mathbf{50}$              & \bf{55.435\% (3.6$\times$)} & \bf{35.870\% (1.1$\times$)}    \vspace{+0.1cm} \\ \hline \vspace{-0.3cm}\\
Null      &     15.217\%     & 31.522\%       \\ \bottomrule
\end{tabular}
\vspace{+0.5cm}
\caption{\textbf{MOA and batch accuracies for different values of the hyperparameter $\mathbold{\lambda}$.} Comparison between a nearest neighbour classifier for MOA and batch predictions against what is expected given random chance. The first line ($\lambda = 0$) denotes our results before adversarial training. By increasing $\lambda$, it is possible to learn ``fairer'' unsupervised representations, invariant to the batch effect confounder, whilst preserving meaningful biological codings. The table also displays fold change (FC) ratios with respect to random performance.}
\label{tab:lambdas}
\end{table}

\section{Discussion}

We presented an approach capable of learning unbiased representations that are biologically meaningful and invariant to uninformative factors of variation. Data representations learned by supervised approaches are intrinsically constrained to the set of cellular phenotypes annotated for training. Our method, based on the unsupervised learning paradigm, does not suffer from such a drawback, and also supports the identification of novel (\textit{i.e.} currently unknown) phenotypes.

Contrary to previous adversarial approaches focusing on supervised learning \cite{louppe,zhang,controllable_invar}, we demonstrated the ability to learn unsupervised representations void of nuisance knowledge whilst retaining as much biological information as possible. This is fundamental across all domains of machine learning where, in order to guarantee good generalisation and unbiased predictions, it is crucial to learn data representations insensitive to known confounders. In particular, this is of paramount importance for machine learning deployment in health care. Identifying and mitigating biases is crucial in order to avoid developing models that mirror human prejudice and/or learn confounds present in data \cite{char2018implementing}. 

We have shown it is possible to remove nuisance variation encoded in a categorical variable (assuming 10 distinct values) using an adversarial approach. However, this method can be trivially generalised to continuous factors by using an adversarial regressor (Figure \ref{architecture}). We are currently analysing an additional imaging data set, derived from the Genotype Tissue Expression (GTEx) Consortium \cite{lonsdale2013genotype}. This resource is comprised of histological slides capturing a range of phenotypes across 53 human tissues, in approximately 800 individuals. Using our approach to learn CAE codings, we have discovered the presence of a continuous confounder (ischemic time), which is correlated with several dimensions of our unsupervised representations (Pearson's $r=0.18$, $p=4.9\times10^{-150}$). We are currently implementing our adversarial learning protocol to remove the ischemic time confounder.

\section{Code and Data Availability}

The BBBC021v1 image set used in this work is freely available from the Broad bioimage benchmark collection web server (\url{https://data.broadinstitute.org/bbbc/}). Furthermore, this paper comes with a dedicated GitHub repository (\url{https://github.com/Nellaker-group/FairUnsupervisedRepresentations}) where we have deposited all the scripts used for the analyses, and the weights of our CAEs. This work was implemented using Python 3 and the deep learning framework PyTorch \cite{paszke2017automatic}. 

\section{Acknowledgements}

We gratefully acknowledge the support of NVIDIA Corporation with the donation of 2 Titan Xp GPUs used for this research. MF is supported through an MRC methodology research grant (MR/M01326X/1). CN is funded through an MRC methodology research fellowship (MR/M014568/1). CML is supported by the Li Ka Shing Foundation, WT-SSI/John Fell funds and by the NIHR Biomedical Research Centre, Oxford, by Widenlife and NIH (5P50HD028138-27).

\newpage

\printbibliography

\beginsupplement

\newpage

\section*{Supplementary Materials}

\vspace{+1cm}

\subsection*{Authors' Affiliations}

Cecilia M. Lindgren is affiliated with: Li Ka Shing Centre for Health Information and Discovery, University of Oxford; Wellcome Trust Centre for Human Genetics, University of Oxford; Program in Medical and Population Genetics, Broad Institute, Cambridge, Massachusetts, USA; Big Data Institute, University of Oxford. 

Christoffer Nell\aa ker is affiliated with: Nuffield Department of Women's \& Reproductive Health, University of Oxford; Big Data Institute, University of Oxford.\\\newline\\\newline\\\newline

\tikzstyle{block} = [draw, rectangle, minimum height=4.5em, minimum width=3em]
\tikzstyle{virtual} = [coordinate]

\begin{figure}[ht!]
\centering
\includegraphics[scale=0.35]{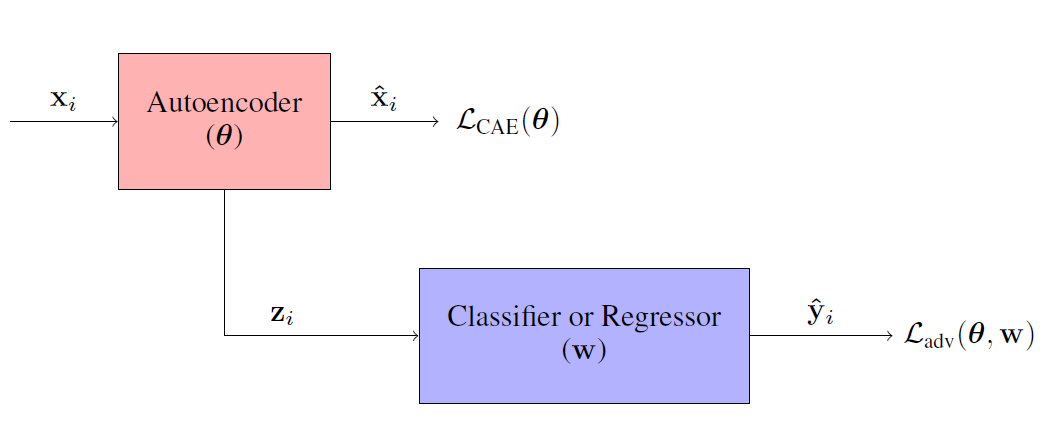}
\centering
\caption{\textbf{Architecture used for adversarial learning.} Our model is comprised of two agents. During training, the convolutional autoencoder is learning compressed representations ($\mathbf{z}_{i}$) of input images, whilst the adversarial neural network (blue block) is removing nuisance knowledge from the CAE codings. The implementation of a classifier (regressor) is dependent on the categorical (continuous) nature of the confounder.}
\label{architecture}
\end{figure}

\FloatBarrier

\end{document}